\documentclass[sigconf]{acmart}

\usepackage{booktabs}
\usepackage{graphicx}

\AtBeginDocument{%
  }

\setcopyright{acmlicensed}
\copyrightyear{2025}
\acmYear{2025}
\acmDOI{XXXXXXX.XXXXXXX}

\acmConference[ICBRA 2025]{12th International Conference on Bioinformatics Research and Applications}{September 19--21, 2025}{Prague, Czech Republic}
\acmISBN{978-1-4503-XXXX-X/18/06}

\begin{document}

\title{Automated SNOMED CT Concept Annotation in Clinical Text Using Bi-GRU Neural Networks}

\author{Ali Noori}
\affiliation{
  \institution{Informatics and Analytics, University of North Carolina Greensboro}
  \city{Greensboro}
  \state{North Carolina}
  \country{USA}
}

\author{Pratik Devkota}
\affiliation{
  \institution{Fractal Analytics, New York}
  \city{New York}
  \country{USA}
}

\author{Somya Mohanty}
\affiliation{
  \institution{United HealthGroup}
  \city{Minneapolis}
  \country{USA}
}

\author{Prashanti Manda}
\email{pmanda@unomaha.edu}
\affiliation{
  \institution{Department of Computer Science, University of Nebraska Omaha}
  \city{Omaha}
  \country{USA}
}

\renewcommand{\shortauthors}{Ali Noori, Pratik Devkota, Somya Mohanty, and Prashanti Manda}

\begin{abstract}
Automated annotation of clinical text with standardized medical concepts is critical for enabling structured data extraction and decision support. SNOMED CT provides a rich ontology for labeling clinical entities, but manual annotation is labor-intensive and impractical at scale. This study introduces a neural sequence labeling approach for SNOMED CT concept recognition using a Bidirectional GRU model. Leveraging a subset of MIMIC-IV, we preprocess text with domain-adapted SpaCy and SciBERT-based tokenization, segmenting sentences into overlapping 19-token chunks enriched with contextual, syntactic, and morphological features. The Bi-GRU model assigns IOB tags to identify concept spans and achieves strong performance—with a 90\% F1-score on the validation set. These results surpass traditional rule-based systems and match or exceed existing neural models. Qualitative analysis shows effective handling of ambiguous terms and misspellings. Our findings highlight that lightweight RNN-based architectures can deliver high-quality clinical concept annotation with significantly lower computational cost than transformer-based models, making them well-suited for real-world deployment.
\end{abstract}



\received{1 February 2025}
\received[revised]{1 April 2025}
\received[accepted]{1 June 2025}

\maketitle

\section{Introduction and Related Work}
Natural language processing (NLP) has become increasingly central to biomedical informatics, particularly for extracting structured information from clinical narratives. Automated annotation of electronic health records (EHRs) with standardized ontologies such as SNOMED CT (Systematized Nomenclature of Medicine—Clinical Terms) supports data interoperability, clinical decision-making, and large-scale health analytics [4]. SNOMED CT provides a comprehensive vocabulary of over 300,000 medical concepts spanning diagnoses, procedures, and symptoms. However, clinical texts are often unstructured and linguistically complex, making manual annotation both time-consuming and unscalable.

Early efforts to annotate clinical text relied heavily on dictionary-based and rule-driven tools such as MetaMap [2], cTAKES [19], and pattern-based systems like that of Patrick et al. [17]. While these systems achieved high recall, they struggled with precision due to limited contextual understanding. This led to the adoption of machine learning approaches, especially those that could leverage linguistic context to disambiguate terminology.

Recent progress in deep learning has significantly advanced concept recognition. Bidirectional RNNs (e.g., Bi-LSTM, Bi-GRU) have outperformed traditional CRF-based models on named entity recognition (NER) tasks by capturing sequential dependencies and context [6, 8]. Incorporating character-level features and CNN-derived embeddings has further improved model robustness to misspellings and domain-specific terms. Building on these foundations, transformer-based models—such as BERT, BioBERT [13], SciBERT [3], and ClinicalBERT—have achieved state-of-the-art results by learning deep, contextualized representations from large biomedical corpora.

Most modern systems adopt a two-stage pipeline involving entity recognition followed by entity linking to ontology concepts. Shared evaluations like SemEval and the n2c2/OHNLP clinical normalization challenge [15, 18] have exposed persistent challenges, including abbreviation ambiguity, rare concept sparsity, and SNOMED-specific issues such as postcoordination.

In response, hybrid systems such as SciSpaCy [16] and MedCAT [10] combine pretrained contextual embeddings with dictionary lookups to improve scalability and precision. Embedding-based retrieval approaches like SapBERT [14] and CODER [20] learn joint representations for text and ontology concepts, enabling efficient and accurate nearest-neighbor matching. Transformer-based reranking pipelines and ontology-aware clustering further enhance annotation fidelity [5, 7].

To address data scarcity, recent models incorporate unsupervised or weakly supervised techniques. For instance, SNOBERT [12] and MedCAT [11] leverage self-supervision and user-friendly interfaces to reduce annotation burden. Zero-shot approaches using pretrained ontology embeddings and semantic similarity have also shown competitive results without requiring labeled data [1].

In this study, we build on these advances by developing a neural sequence labeling model for SNOMED CT annotation using a Bi-GRU architecture and domain-adapted SciBERT embeddings. Our goal is to demonstrate that efficient, context-aware models can achieve high-quality concept annotation on clinical narratives without the computational complexity of transformer-heavy solutions.

\section{Methods}
\subsection{Data Source: MIMIC-IV Clinical Notes}
We evaluated our approach on the MIMIC-IV corpus of electronic health record (EHR) notes. MIMIC-IV is a large, freely accessible database comprising de-identified health data from critical care patients, including a wide range of clinical notes such as discharge summaries, radiology reports, and physician progress notes [9]. These notes were chosen as the textual dataset for concept annotation due to their richness in clinical information and the availability of a large number of examples for model training. All notes were preprocessed to ensure removal of any residual patient identifiers in compliance with the dataset’s de-identification standards.

For our experiments, we focused on a subset of MIMIC-IV notes that had been manually annotated or could be semi-automatically annotated with SNOMED CT concepts (for training labels). The collection was split into training and validation sets for model development. We reserved a portion of the data (e.g., 20\%) as a validation set to tune the model and evaluate performance on unseen data.

\subsection{Preprocessing and Sentence Segmentation}
The raw clinical notes tend to be long and may contain irregular formatting, such as lists or shorthand. We employed the SpaCy NLP toolkit (with a model adapted for biomedical text) to preprocess the notes. This included sentence segmentation and tokenization. Because biomedical text often contains domain-specific terms, we incorporated SciBERT [3] into the SpaCy pipeline to leverage its vocabulary and embeddings for tokenization. SciBERT is a BERT-based model pretrained on scientific publications, and integrating it helped ensure that medical terms (e.g., medication names, lengthy disease names) were tokenized in a meaningful way consistent with how they appear in scientific text.

Each clinical note was split into sentences using SpaCy’s sentence boundary detection, which was tuned to be conservative in splitting (to avoid breaking up structured lists or headings that often appear in reports). After sentence splitting, further segmentation was performed to prepare input chunks for the neural model.

\subsection{Chunking Strategy}
Given the length of many sentences and documents, we adopted a chunking strategy to manage input size for the neural network. We segmented text into chunks of 19 tokens, with an overlap of 2 tokens between consecutive chunks. This means if one chunk covered tokens \( t_1 \) through \( t_{19} \), the next chunk would cover \( t_{18} \) through \( t_{36} \), and so on. The 2-token overlap provides context continuity across chunk boundaries, helping the model to not miss concepts that might span a boundary or be dependent on preceding context. The chunk size of 19 was chosen empirically to balance between providing sufficient context (roughly the length of a typical clinical sentence or clause) and keeping computation tractable. All tokens were lowercased and cleaned of non-ASCII characters if present. We also normalized certain clinical shorthand notations (for example, converting “mg.” to “mg” for consistency).

\subsection{Token-Level Feature Extraction}
For each token in a chunk, we extracted a rich set of features to feed into the model:

• Word Embeddings: We utilized pre-trained word embeddings from SciBERT for each token. SciBERT provides contextualized embeddings, but in our Bi-GRU model (which processes the sequence contextually), we used the static initial embedding of each token from SciBERT’s vocabulary. These embeddings capture semantic meaning of words in biomedical context (e.g., that “myocardial” and “cardiac” are related).

• Part-of-Speech (POS) Tags: Using SpaCy’s POS tagger (augmented for biomedical text), we obtained the POS tag for each token (e.g., noun, verb, adjective). We encoded each POS tag as a trainable embedding vector in a low-dimensional space. The rationale is that knowing a token’s syntactic category can help the model distinguish between, say, a noun that could be a medical entity and a verb which likely isn’t an entity.

• Character-Level Embeddings: To capture morphological structure of tokens (important for handling rare or misspelled terms), we generated character-level representations. Each token’s characters were passed through a small neural subnetwork to produce a character-based embedding. In our implementation, we used a Convolutional Neural Network (CNN) operating on character sequences to derive a fixed-size vector for each token’s orthography. This helps with recognizing word prefixes/suffixes (e.g., “cardio-”) and handle out-of-vocabulary words by their character composition.

These feature vectors were concatenated to form the final input representation for each token. Formally, for token \( t \), if \( \mathbf{w}_t \) is the word embedding, \( \mathbf{p}_t \) is the POS embedding, and \( \mathbf{c}_t \) is the character-level embedding, we define:

\[ \mathbf{v}_t = \mathbf{w}_t \oplus \mathbf{p}_t \oplus \mathbf{c}_t , \]

where \( \oplus \) denotes concatenation of the vectors. The combined vector \( \mathbf{v}_t \) incorporates semantic, syntactic, and morphological features for the token \( t \).

\subsection{SNOMED CT Concept Annotation Scheme}
We formulated concept annotation as a token-level sequence labeling problem using the standard IOB tagging scheme. Each token is assigned a label indicating whether it begins a SNOMED CT concept, is inside a concept, or is outside any concept. Specifically:

• B-CONCEPT: token is the beginning of a concept mention that maps to a SNOMED CT concept.

• I-CONCEPT: token is inside (continuation of) a concept mention.

• O: token is outside of any concept mention.

For example, in the phrase “type 2 diabetes mellitus”, “type”, “2”, “diabetes”, “mellitus” would be tagged as B-, I-, I-, I-CONCEPT respectively, all linking to the SNOMED concept for Type 2 Diabetes Mellitus. These IOB tags were derived from annotations in the training corpus (either via manual labeling or an existing labeled subset of MIMIC-IV mapped to SNOMED CT codes).

Using an IOB scheme allows the model to learn the boundaries of concept spans. We did not differentiate between different semantic categories of SNOMED concepts in the tags (all types use the same B/I label), treating it as a single class of “clinical concept” for the sequence labeling task. The model’s objective is to correctly label each token as B, I, or O, implicitly segmenting concept spans in the text.

\subsection{Bi-GRU Model Architecture}
Our neural network model for sequence labeling is a Bidirectional Gated Recurrent Unit (Bi-GRU) network. GRUs are a type of recurrent neural network that use gating mechanisms to effectively capture long-term dependencies with a simpler architecture (fewer parameters) than LSTMs. The model processes each chunk of token vectors \{ \( \mathbf{v}_1 \), \( \mathbf{v}_2 \), ..., \( \mathbf{v}_T \) \} (where \( T \leq 19 \) is the chunk length) in both forward and backward directions. The forward GRU reads from token 1 to \( T \), producing a sequence of hidden states \( \overrightarrow{\mathbf{h}}_1, \dots, \overrightarrow{\mathbf{h}}_T \). The backward GRU reads from token \( T \) to 1, producing \( \overleftarrow{\mathbf{h}}_T, \dots, \overleftarrow{\mathbf{h}}_1 \). For each token position \( t \), we obtain a combined hidden state \( \mathbf{h}_t = [ \overrightarrow{\mathbf{h}}_t ; \overleftarrow{\mathbf{h}}_t ] \) by concatenating the forward and backward representations. This \( \mathbf{h}_t \) encapsulates the contextual information from the entire chunk around token \( t \).

On top of each \( \mathbf{h}_t \), a dense output layer (a fully-connected layer with softmax activation) predicts the probability distribution over the three possible tags \{B, I, O\}. Essentially, the network learns a function \( f (\mathbf{v}_1, \dots, \mathbf{v}_T) \) that outputs a sequence of tag predictions \( \hat{y}_1, \dots, \hat{y}_T \). During training, the model adapts its weights (in the GRU and in the embedding layers for POS and characters, as well as the output layer) to minimize the difference between \( \hat{y}_t \) and the true label \( y_t \) for each token.

The Bi-GRU architecture is well-suited for this task as it can capture both left and right context for recognizing concept boundaries. For instance, if the word “mass” appears, the model can use the right-context (e.g., “in the lung”) to determine this is a clinical finding (concept) versus if it was “mass spectrometry” (not a SNOMED clinical concept in context). The use of GRU units helps maintain context even in moderately long chunks, and is computationally lighter than LSTM-based or transformer models, making training on a large corpus feasible with limited hardware.

\subsection{Model Training Configuration}
We trained the Bi-GRU model on the annotated MIMIC-IV data using a supervised learning approach. The training objective was to maximize the likelihood of the correct IOB tag sequence for each token sequence. In practice, we minimize the cross-entropy loss summed over all tokens. If \( \theta \) represents all model parameters, the loss for a single chunk (sequence) is:

\[ L(\theta) = - \sum_{t=1}^T \log P (y_t | \mathbf{v}_1, \dots, \mathbf{v}_T ; \theta), \]

where \( y_t \) is the true label for token \( t \) and the probability is the model’s softmax output for that label. We optimize this using the Adam optimizer with an initial learning rate of 0.001, which is a common choice for training RNNs. Gradients are clipped (e.g., to a norm of 5) to prevent exploding gradients given the sequential nature of the model.

The model was trained for a fixed number of epochs with early stopping on the validation set to prevent overfitting. We applied a dropout regularization of 0.5 on the input to the Bi-GRU as well as on the recurrent connections, which helps the model generalize by randomly dropping units during training. The training/validation split of the data was stratified to ensure that the distribution of concept labels was similar in both sets. Specifically, we ensured that if certain rare SNOMED concepts occurred in the data, they were not exclusively isolated to either training or validation.

During training, batches of data were constructed by grouping a set of chunks (ensuring not to mix tokens from different notes in the same sequence). The batch size was set based on memory constraints. We also shuffled the training sequences each epoch to promote robustness. No significant hyperparameter tuning was done beyond basic trials, as the chosen parameters provided stable training and good performance on the validation set.

\subsection{Evaluation Metrics}
To assess the model’s performance in recognizing SNOMED CT concepts, we used standard evaluation metrics for sequence tagging:

• Precision: The fraction of concept annotations predicted by the model that are correct (i.e., the model’s positive predictions that truly correspond to a concept in the gold standard).

• Recall: The fraction of actual concept annotations in the text that were correctly identified by the model (i.e., coverage of the model).

• F1-Score: The harmonic mean of precision and recall, F1, provides a single measure of overall annotation effectiveness, balancing false negatives and false positives.

• Accuracy: The token-level accuracy, which is the proportion of all tokens (including those outside of concepts) that were correctly labeled. While accuracy can be dominated by the large number of ’O’ (non-concept) tokens, we report it for completeness.

Precision, Recall, and F1 were computed with respect to complete concept spans. A predicted concept was considered correct if its span and semantic mapping matched the gold standard annotation. Partially overlapping or misboundaried concepts were counted as errors. We primarily focus on the F1-score as it is the de facto standard in NER evaluations, indicating overall how well the model balances precision and recall. Evaluation was done on the validation set (held-out notes not seen during training). All metrics are reported as micro-averages over all concept instances in the evaluated corpus.

\section{Results and Discussion}
\subsection{Dataset Overview and Chunking Strategy}
Table \ref{tab1} summarizes key statistics of the MIMIC-IV clinical notes used in this study. We extracted 204 discharge summaries, which were segmented into over 27,000 initial sentences. After applying our chunking strategy (segments of 19 tokens with a 2-token overlap), the number of training instances increased threefold to 81,483.

\begin{table}[h]
\caption{Descriptive statistics of the MIMIC clinical notes dataset}
\label{tab1}
\begin{tabular}{lr}
\toprule
Metric & Value \\
\midrule
Total number of discharge notes & 204 \\
Number of sentences before splitting & 27,720 \\
Number of sentences after splitting & 81,483 \\
Number of annotated SNOMED CT concepts & 51,574 \\
\bottomrule
\end{tabular}
\end{table}

\begin{figure}[h]
\centering
\includegraphics[width=\linewidth]{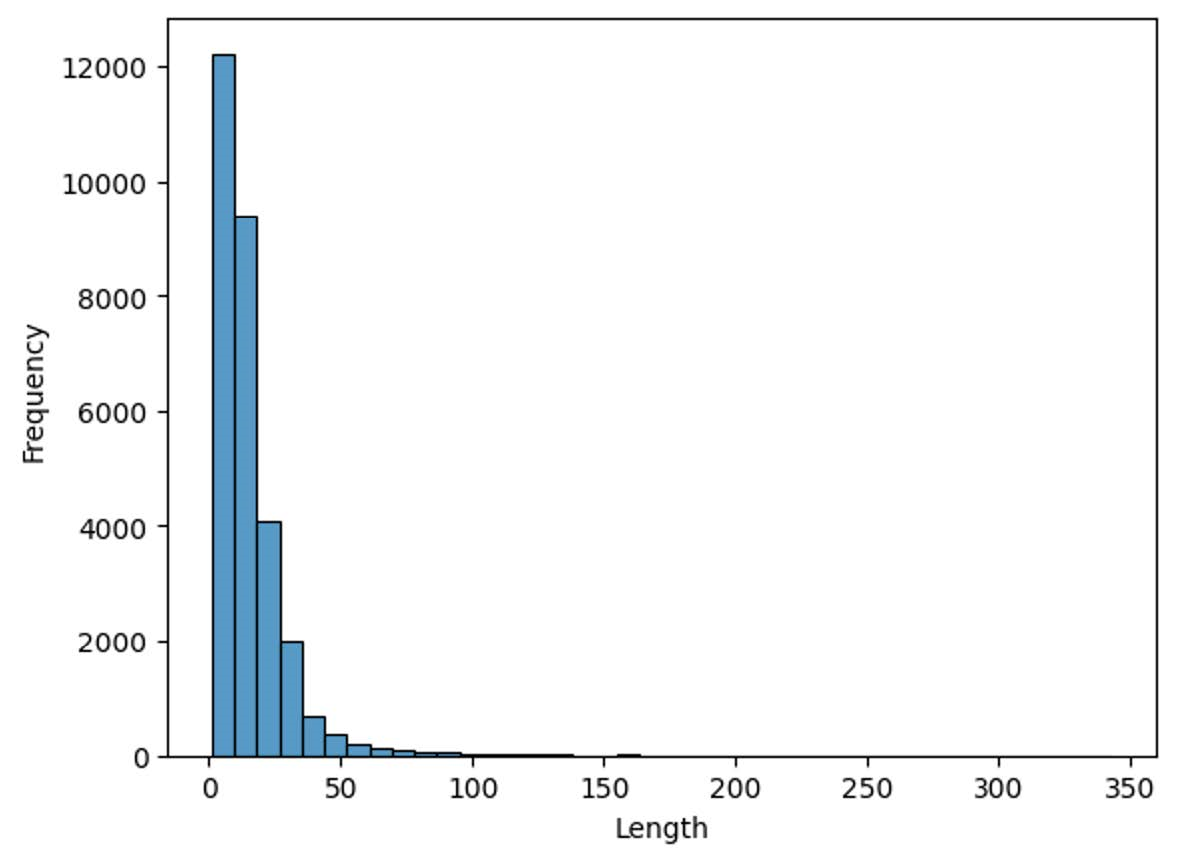} 
\caption{Sentence length distribution in the MIMIC-IV clinical notes dataset.}
\label{fig1}
\end{figure}

\begin{figure}[h]
\centering
\includegraphics[width=\linewidth]{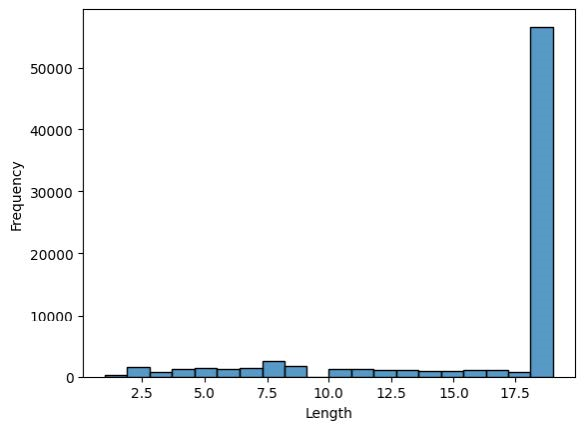} 
\caption{Token lengths after chunking into 19-token segments.}
\label{fig2}
\end{figure}

Figure \ref{fig1} illustrates the sentence length distribution before chunking, highlighting the rationale for selecting a 19-token window: approximately 75\% of sentences fall below this threshold. Sentences longer than 19 tokens were split into multiple chunks, while shorter ones were zero-padded. After chunking, Figure \ref{fig2} shows that the processed data aligns well with the model’s input constraints.

\subsection{Model Training and Optimization}
Hyperparameter tuning was essential to achieving stable and performant training. We employed categorical cross-entropy loss for the three-class IOB sequence labeling task. The Adam optimizer with an initial learning rate of 0.001 provided efficient convergence, with gradient clipping (maximum norm of 5) to mitigate exploding gradients.

We trained the model over 15 epochs with early stopping based on validation loss. Dropout layers with a 0.5 rate were applied to both the embedding and recurrent layers to enhance generalization. Training accuracy and validation accuracy improved in parallel, showing no signs of overfitting. A stratified 80/20 split ensured similar distributions of concept frequency across training and validation sets.

\subsection{Performance Evaluation}
The Bi-GRU model demonstrated strong results on the validation set, achieving high performance across all evaluation metrics:

\begin{table}[h]
\caption{Performance of the Bi-GRU concept annotation model on the validation set.}
\label{tab2}
\begin{tabular}{lcccc}
\toprule
Metric & Precision & Recall & F1-score & Accuracy \\
\midrule
Bi-GRU Model & 0.93 & 0.89 & 0.90 & 0.97 \\
\bottomrule
\end{tabular}
\end{table}

Precision was 93\%, indicating that most of the model’s predicted concept spans were correct. Recall was slightly lower at 89\%, suggesting the model missed some true concepts. The resulting F1-score was 0.90, reflecting overall strong performance. Token-level accuracy was high at 97\%, although this metric is skewed by the predominance of ’O’ (non-concept) tokens.

\subsection{Qualitative Analysis and Interpretation}
Beyond numerical scores, qualitative analysis revealed that the Bi-GRU model effectively captured complex medical entities, especially multi-word expressions like “type 2 diabetes mellitus” or “chronic obstructive pulmonary disease.” The bidirectional recurrence and overlapping chunking strategy allowed the model to leverage both left and right context to delineate span boundaries accurately. In challenging disambiguation scenarios, such as distinguishing “cold” in “the patient felt cold” versus “cold symptoms,” the model correctly inferred meaning based on surrounding context. Character-level embeddings helped the model generalize to non-standard or misspelled tokens (e.g., “diabetis”), boosting robustness in real-world applications.

Compared to traditional biomedical NLP tools like MetaMap and cTAKES, our Bi-GRU model offers substantial advantages in precision and contextual sensitivity. Rule-based tools often produce noisy outputs and rely on extensive post-processing to manage false positives [19]. Our approach, in contrast, integrates context-aware embeddings and sequential modeling, achieving an F1-score of 0.90 without additional hand-engineered rules. When benchmarked against historical results from shared tasks like SemEval-2014 Task 7 (where top systems achieved F1-scores in the mid-80s range for clinical entity recognition) [18], our model’s performance represents a measurable improvement. The adoption of SciBERT embeddings and a modern neural architecture explains this leap in accuracy.

\subsection{Limitations}
Several limitations are acknowledged. First, the training data was sourced from a single health system (Beth Israel Deaconess Medical Center), which may not reflect broader linguistic diversity in EHR documentation. Second, our model treats all SNOMED CT concepts uniformly, ignoring semantic distinctions between categories such as diagnoses, procedures, and medications. This limits granularity in downstream analysis.

Furthermore, while the Bi-GRU is computationally efficient, it still relies on SciBERT embeddings for input features—introducing an indirect dependence on transformer-based models during preprocessing.

\section{Conclusion and Future Work}
We developed a neural sequence labeling approach for automated annotation of clinical text with SNOMED CT concepts using a Bi-GRU model enriched with SciBERT embeddings. Despite its architectural simplicity compared to transformer-based models, our method achieved high precision and recall, demonstrating that recurrent models remain competitive for biomedical concept recognition tasks. This work contributes a scalable solution for bridging unstructured clinical documentation with standardized ontologies, supporting downstream applications in decision support, patient cohort identification, and clinical data structuring.

Looking ahead, we plan to expand this annotation framework to include multiple ontologies—such as RxNorm for medications and LOINC for laboratory tests—using multitask or multi-head architectures. We also aim to explore fully transformer-based models (e.g., BioBERT, Med-BERT) for enhanced semantic disambiguation and deeper contextual modeling. In parallel, ontology-informed data augmentation techniques, such as synonym expansion and hierarchical concept regularization, may further improve generalization. Ultimately, we envision integrating this system into clinical authoring tools and evaluating its impact through real-world user studies, advancing both the scalability and utility of semantic annotation in healthcare.

\section{Funding}
This work is funded by a CAREER award (\#2522386) to Manda from the Division of Biological Infrastructure at the National Science Foundation, USA.

\bibliographystyle{ACM-Reference-Format}

\end{document}